\pdfoutput=1

\documentclass[11pt]{article}

\usepackage[]{acl2023}
\usepackage{times}
\usepackage{latexsym}
\usepackage{diagbox}
\usepackage{tabularx}
\usepackage{multirow}
\usepackage{subcaption}
\usepackage{algorithm}
\usepackage{algorithmic}
\usepackage{subcaption}

\usepackage{times}
\usepackage{latexsym}
\usepackage{graphicx}
\usepackage{amsmath}
\usepackage{amssymb}
\usepackage{booktabs}
\usepackage{adjustbox}
\usepackage{geometry}

\usepackage{arydshln}
\usepackage{xcolor}

\usepackage[T1]{fontenc}

\usepackage[utf8]{inputenc}

\usepackage{microtype}
\usepackage[T1]{fontenc}  
\usepackage{hyperref}    
\usepackage{url}      
\usepackage{booktabs}    
\usepackage{amsfonts}    
\usepackage{nicefrac}    
\usepackage{microtype}   
\usepackage{xcolor}     

\usepackage[normalem]{ulem}

\usepackage[utf8]{inputenc}

\usepackage{microtype}

\usepackage{inconsolata}

\title{Learning to Diversify Neural Text Generation via
\\ \textit{De}generative Model}

\author{
        Jimin Hong\thanks{\hspace{0.2cm} Equal contribution} 
        \And ChaeHun Park\footnotemark[1] \\
        KAIST AI \\
        \tt{\{jimmy.h, ddehun, jchoo\}@kaist.ac.kr} \\  
        \And Jaegul Choo \\
}

\begin{document}
\maketitle
\begin{abstract}

Neural language models often fail to generate diverse and informative texts, limiting their applicability in real-world problems. 
While previous approaches have proposed to address these issues by identifying and penalizing undesirable behaviors (e.g., repetition, overuse of frequent words) from language models,
we propose an alternative approach based on an observation: models primarily learn attributes within examples that are likely to cause \textit{degeneration} problems.
Based on this observation, we propose a new approach to prevent degeneration problems by training two models.
Specifically, we first train a model that is designed to amplify undesirable patterns.
We then enhance the diversity of the second model by focusing on patterns that the first model fails to learn.
Extensive experiments on two tasks, namely language modeling and dialogue generation, demonstrate the effectiveness of our approach.

\end{abstract}

\section{Introduction}

Neural text generation is a fundamental task including open-ended applications such as language modeling or dialogue generation~\cite{chen2017survey}.
Despite considerable advances in the task, generation models often result in \textit{degeneration}~\cite{li2016diversity, dinan2019second, holtzman2019curious} such as repetition or the overproduction of dull and generic texts with lack of diversity.

Previous studies have proposed to overcome these issues as follows:
\citet{welleck2019neural} suggests to explicitly penalize repetition using unlikelihood objective.
\citet{li2019don} applies unlikelihood training~\cite{welleck2019neural} to dialogue domain by penalizing overuse of common words in generated responses.
\citet{jiang2019improving} and \citet{choi2020f} refine the Maximum Likelihood Estimation~(MLE) objective by considering the frequency distribution of words.
In other words, prior works focus on explicitly defining undesirable behaviors and penalizing them in a training phase.
Although these studies have shown promising results, we argue that identifying such negative behaviors of models can be laborious and task-dependent.

Instead, we propose a novel approach that does not require explicitly specifying the negative behaviors of generation models.
Our approach is based on a fundamental observation (§\ref{sec::preliminary}): 
Models are misguided by attributes within training examples that may be harmful to reflecting human diversity.
Based on the observation, we propose \textsc{LfD}: \textbf{L}earning \textbf{f}rom \textbf{D}egeneration, a novel approach to remedy degeneration problems in open-ended applications.
Specifically, we first train a model which is designed to \textit{De}generate by amplifying undesirable patterns in examples (§\ref{sec::degenerative}).
We then train the second model to enhance its diversity by leveraging the predictions of the first model (§\ref{sec::poe}).
Experimental results on two representative open-ended generation tasks demonstrate the effectiveness of our approach.

In summary, our contributions include: 
\begin{itemize}
    \setlength\itemsep{0.01em}
\item We analyze how the learning dynamics of training examples are affected based on the degree of their diversity on open-ended text generation tasks.

\item We propose a novel approach that enhances the overall generation quality, especially diversity.

\item \textsc{LfD} can be easily applied regardless of tasks in open-ended applications. 

\end{itemize}

\section{Related Work}
Recent studies have reported that neural generation models often make various forms of degeneration problems~\cite{li2016diversity,holtzman2018learning,dinan2020second}.
Several methods have suggested training objectives to remedy this problem by alleviating token distribution mismatch between human and machine-written texts~\cite{jiang2019improving}, balancing token distribution~\cite{choi2020f}, or directly penalizing negative behaviors on generated texts with auxiliary loss~\cite{welleck2019neural,he2019negative}.
\citet{wang2021diversifying} address over-confidence issues in text generation by adaptive label smoothing.
\citet{li2022diversifying} leverages a task-specific data filtering process~\citet{csaky2019improving} to build negative teacher for dialogue generation.
Such studies are orthogonal to \textsc{LfD} since we mainly focus on training dynamics of examples that are available regardless of tasks.

\section{Preliminary Study}
\label{sec::preliminary}
Previous studies have reported that the generation quality is likely to be degraded due to inherent attributes within the training examples, such as token repetition~\cite{welleck2019neural,fu2021theoretical}, a skewed frequency distribution of words~\cite{fagan2011introduction}, and genericness in responses~\cite{csaky2019improving}.
We refer to such attributes as \textit{degenerative attributes} in the paper.
In this section, we analyze how such \textit{degenerative attributes} affect the learning dynamics of training examples.
We conduct experiments on two open-ended text generation tasks: language modeling and dialogue generation.

\subsection{Setup}
\label{sec::prelimiary_setup}
\noindent\textbf{Dataset}\indent
For language modeling, we use WikiText-103~\cite{merity2016pointer}, a collection of English documents extracted from verified Wikipedia.
For dialogue generation, we use DailyDialog~\cite{li2017dailydialog} consisting of open-domain dialogues that reflect daily conversations. 

\noindent\textbf{Metrics}\indent
We use the following metrics to measure the \textit{degenerative attributes} in each example. 
For language modeling, we use \textbf{Average Frequency} to evaluate the lexical diversity of each example by averaging the frequency of tokens in an example.
We also leverage \textbf{Repetition}~\cite{welleck2019neural} that measures how often each token already appears in the previous part of an example.

For dialogue generation, we regard \textbf{Source Entropy}~\cite{csaky2019improving} as the measurement of 
how trivial response is.
A response with higher entropy indicates to correspond with more dialogue histories.
We also use \textbf{Context Overlap}~\cite{li2019don} that calculates the bi-gram overlap between dialogue histories and responses.
We describe further details of each metric in Appendix~\ref{appendix::metrics_preliminary}.

\begin{figure}[t!]

\begin{subfigure}{0.48\textwidth}
\includegraphics[width=.95\linewidth]{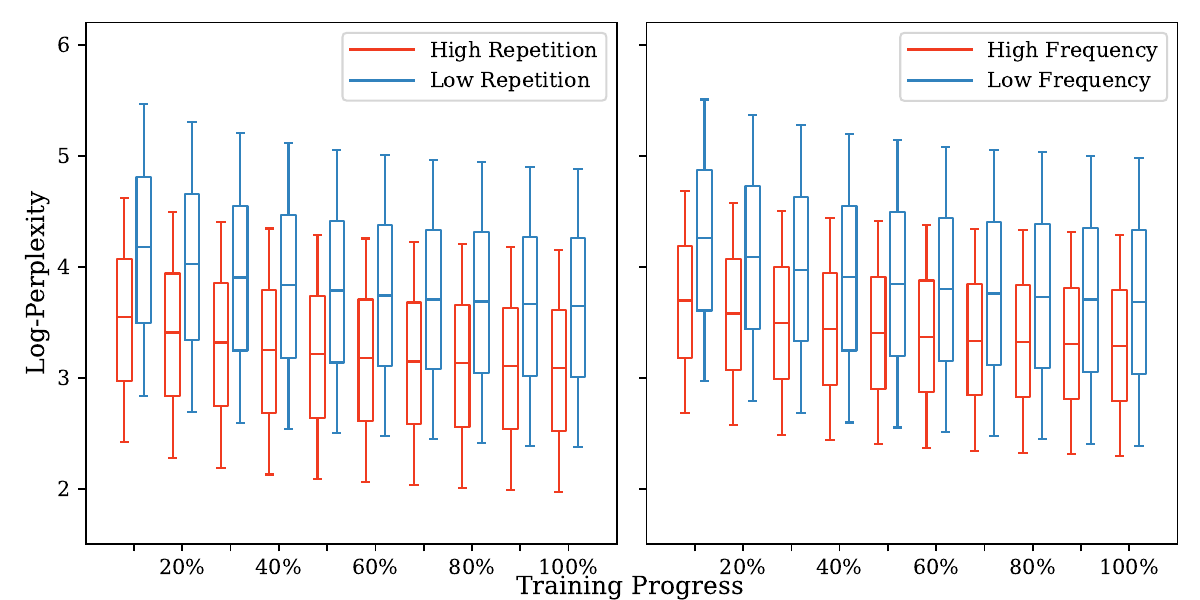}
\caption{WikiText-103}
\end{subfigure}

\begin{subfigure}{0.48\textwidth}
\includegraphics[width=.95\linewidth]{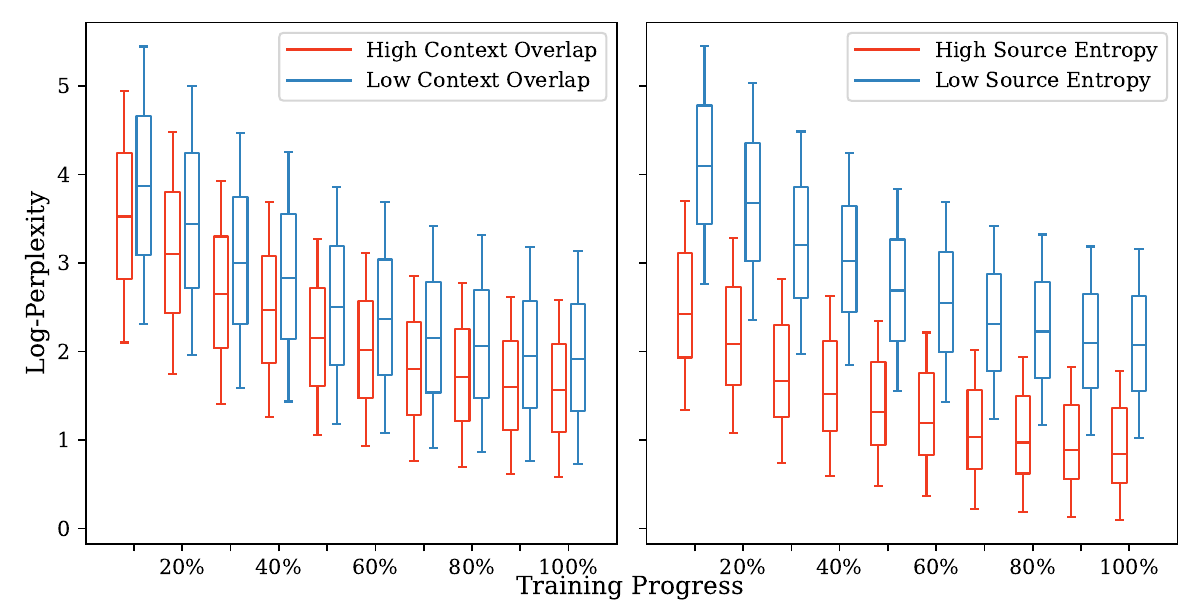}
\caption{DailyDialog}
\end{subfigure}
\caption{Comparisons between two groups with \textcolor{red}{high} and \textcolor{blue}{low}  degenerative attributes on language modeling~(top) and dialogue generation~(bottom) tasks. 
}
\label{fig:training_dynamics}
\vspace{-0.3cm}
\end{figure}

\noindent\textbf{Model}\indent
We train 6-layer transformer~\cite{vaswani2017attention} decoder and 6-layer transformer encoder-decoder models from scratch for language modeling and dialogue generation tasks, respectively.
To analyze the training dynamics of models on each task, we first train models and save their checkpoints at each epoch.
We then divide training examples into two groups based on the attribute score\footnote{We choose top and bottom of 5k examples with high and low attribute score, respectively.}, and compute log-perplexity at each stage during training.\footnote{For WikiText-103, we compute the average log-perplexity at sentence-level.} 

\subsection{Analysis}
\label{sec::prelimianry_result}
Figure~\ref{fig:training_dynamics} shows log-perplexity of examples with low and high degenerative attributes according to training progress.
Specifically, the group with high degenerative attribute usually have lower perplexity than the other group.
Even though the perplexity of examples in low degenerative attribute monotonically decreases as training progresses, it does not imply that the model generates diverse sequences in test time since examples with high attributes are still more likely to be produced than others.

\section{\textsc{LfD}: Learning from \textit{De}generation}
Based on the analysis~(§\ref{sec::preliminary}), we argue that the model should be prevented from overfitting to degenerative attributes.
Inspired by previous studies~\cite{nam2020learning,sanh2021learning},
we propose a new training approach consisting of two steps:
(a) intentionally train a model $f_{\theta_{D}}$ to amplify degenerative attributes in examples, and (b) train a diversity-enhanced model $f_{\theta_{M}}$ by leveraging $f_{\theta_{D}}$.

\subsection{Background}
Generative language model $f_{\theta}$ is usually trained
to maximize conditional probability distribution of $\textbf{p}(\textbf{y}|\textbf{x},\theta)$,
where $\textbf{x} = (x_1,\ldots, x_{|\textbf{x}|})$ and $\textbf{y} = (y_1,\ldots, y_{|\textbf{y}|})$ are input and target sequences.
A typical approach to train the model is optimizing $\theta$ by minimizing the following negative log-likelihood:  
\begin{equation}
\mathcal{L}_{\text{MLE}}(\theta,\textbf{x},\textbf{y}) =  - \sum_{t=1}^{\textbf{|y|}} \log p(y_{t}|y_{<t},\textbf{x},\theta)
\label{eq::mle}
\end{equation}
\subsection{Training \textit{De}generative Model $f_{\theta_{D}}$}
\label{sec::degenerative}
From the analysis~(§\ref{sec::preliminary}), we observe that the difference between the two groups of examples is significant, especially in the early training phase.
We enforce \textit{De}generative model $f_{\theta_{D}}$ to overfit degeneration attributes captured in a small number of iteration.
In particular, we leverage the truncated cross entropy loss~\cite{han2018co} to amplify attributes at token-level.
The procedure is following: (a) we train $f_{\theta_{D}}$ by $K$ step with standard training. 
(b) After $K$ steps, only $R$\% of tokens with small loss within a batch are used to update the model $f_{\theta_{D}}$ by $H$ steps.
We expect that tokens are potentially generic when $f_{\theta_{D}}$ predicts them with high confidence.
Conversely, tokens are potentially diverse when $f_{\theta_{D}}$ predicts them with low confidence.

\subsection{Enhancing Diversity via 
\textit{De}generative model $f_{\theta_{D}}$}
\label{sec::poe}
Now we explain how to encourage the diversity of the main model $f_{\theta_{M}}$ by exploiting the predictions of $f_{\theta_{D}}$.
Inspired by \citet{utama2020towards}, we introduce Product-of-Expert (PoE) to prevent $f_{\theta_{M}}$ from learning degenerative attributes amplified in $f_{\theta_{D}}$.
Namely, the model $f_{\theta_{M}}$ is likely to concentrate on attributes that $f_{\theta_{D}}$ fails to learn.
Specifically, the model $f_{\theta_{M}}$ is trained with the predictions of $f_{\theta_{D}}$ by combining their outputs as: 
\begin{equation}
\begin{aligned}
& \sigma_{\text{poe}}(\theta_D,\theta_M,\textbf{x},\textbf{y},t) \\
& = \log p({y_t|y_{<t},\textbf{x}},\theta_{D}) +\log p({y_t|y_{<t},\textbf{x},\theta_{M}})
\end{aligned}    
\end{equation}

Combined predictions are used to calculate the loss for model optimization.
During the training, we only optimize the parameters of $f_{\theta_{M}}$ while keeping the parameter of $f_{\theta_{D}}$ as frozen. 
\begin{equation}
\begin{aligned}
\vspace{-0.2cm}
& \mathcal{L}_{\text{PoE}}(\theta_{M},\textbf{x},\textbf{y})\\ 
&= -\sum_{t=1}^{\textbf{|y|}} \log \text{softmax} (\sigma_{\text{poe}}(\theta_D,\theta_M,\textbf{x},\textbf{y},t))
\end{aligned}    
\end{equation}
The final loss is combined as  $\mathcal{L}_{\text{MLE}}$ + $\lambda\mathcal{L}_{\text{PoE}}$. 
In test time, we generate sequences using $f_{\theta_{M}}$ only.


\begin{table}[t]

\resizebox{0.99\linewidth}{!}{
\begin{tabular}{c c l l c r}
\toprule[0.3ex]

\textbf{Model}
 &\textbf{PPL} &\textbf{KLD} &  \textbf{ZipC}  & \textbf{Rep.} & \textbf{Uniq.}  \\
\toprule
\textbf{MLE}    
&\textbf{26.3}   &2.3 & 1.16     & 1.3 & 6.0k  \\
\textbf{$\text{UL}^{\dagger}$}
&26.9 &\underline{2.1}  & \underline{1.06}      & \textbf{0.7} & \underline{7.2k}   \\

\textbf{Focal} 
&\underline{26.7} &2.3 & 1.15     & 1.4 & 5.9k      \\
\textbf{LfD} 
&26.9 &\textbf{1.9} &   \bf{0.94}    & \underline{0.8} & \bf{8.4k}    \\
\textbf{Human}  
&-      &-      & 0.93  & 0.2 & 10.9k \\ 

\hdashline
\noalign{\vskip 0.75ex}
$f_{\theta_{D}}$ 
& 118.15 &2.9 &1.20  & 3.6 & 4.1k    \\
\textbf{LfD}$_{\text{MLE}}$ 
& 26.7 & 2.0 &1.09   & 1.4 & 6.4k    \\
\toprule
\end{tabular}
}
\caption{Evaluation results on WikiText-103. Top-$k$ sampling~\cite{fan2018hierarchical} is selected as decoding algorithm with $k$=20.
We attach $\dagger$ to baselines that explicitly penalize negative behavior (e.g. repetition or frequency).
The best and the second best results
are highlighted in \textbf{bold} and \underline{underline}, respectively. 
The results close to human gold standard are regarded as better performance.
For \textbf{PPL} and \textbf{KLD}, the lowest scores are best performances.
}

\vspace{-0.3cm}
\label{app:lm_table}
\end{table}

\begin{table*}
\centering
\begin{subtable}{\textwidth}
\centering
\begin{tabular}{cccccccccc}
\toprule[0.3ex]
  \multicolumn{1}{c}{}& \multicolumn{2}{c}{\textbf{BLEU}} &    \multicolumn{2}{c}{\textbf{Distinct}} &\multicolumn{2}{c}{\textbf{self-BLEU}}  & \multicolumn{1}{c}{\textbf{KLD}} & \multicolumn{1}{c}{\textbf{Context}} & \multicolumn{1}{c}{\textbf{Source}} \\ 
\multicolumn{1}{c}{\textbf{Model}} &  $n$=2 & $n$=3  &   $n$=2 & $n$=3  & $n$=3 & $n$=4 & &\textbf{Overlap}&\textbf{Entropy}\\ \toprule

\textbf{MLE}     & 	15.85&	8.67&	   5.48&	15.76&		96.30&	92.08&	   2.23 & 12.42 &0.88\\
\textbf{$\text{Dialogue-UL}^{\dagger}$}& 	19.19&	\underline{12.22}&		12.21&	32.78&	90.47&	80.62&	\underline{1.38} &\underline{10.49} &0.66\\
\textbf{$\text{FACE}^{\dagger}$}    & 	7.44&	3.90&		\underline{14.45}&	\underline{42.89}&	\textbf{85.64}&	\textbf{71.07}&	1.53 & 3.66 &\underline{0.12}\\
\textbf{Focal}   & 15.85&	8.62&	   5.48&	15.72&	    96.31&	92.05&	   2.25 & 12.88 &0.93\\
\textbf{CP}      & 	\underline{19.22}&	12.19&		11.94&	31.31&	90.69&	81.50&	1.40 &10.94 & 0.68\\  
\textbf{LfD}    & 	\textbf{19.26}&	\textbf{12.37}&		\textbf{16.52}&	\textbf{42.92}&		\underline{86.44}&	\underline{73.09}&		\textbf{1.12} & \textbf{9.38}& \textbf{0.55}\\
\textbf{Human}   & 	-    &	-    &		35.97&	67.00& 68.72&	47.34&-&     9.83 & 0.35\\
\cdashline{0-9}\noalign{\vskip 0.75ex}
\textbf{$f_{\theta_{D}}$}     &	13.63	&6.77&	1.15&	2.76&	99.53&  99.02&	4.10&	16.07&	1.13\\
\textbf{LfD}$_{\text{MLE}}$    &	19.59&	12.67&	15.03&	39.04&	88.03&	76.04&	1.16 & 9.85&0.63\\
\bottomrule
\end{tabular}
\end{subtable}

\caption{Evaluation results on DailyDialog. Greedy decoding algorithm is used for all models, following \citet{jiang2019improving}.
The indicators are the same as Table~\ref{app:lm_table}.
For \textbf{BLEU}, we regard the highest scores as the best performances.
}
\label{app:dialogue_table}
\vspace{-0.2cm}
\end{table*}

\section{Experiments}
\subsection{Task}
We evaluate \textsc{LfD} on language modeling, dialogue generation, and abstractive summarization tasks with datasets described in Section~\ref{sec::prelimiary_setup}: WikiText-103~\cite{merity2016pointer}, DailyDialog~\cite{li2017dailydialog}, and CNN/DailyMail~\cite{nallapati2016abstractive}.
Further details of each dataset are in Appendix~\ref{appendix::dataset}.

\subsection{Setup}
\noindent\paragraph{Baselines}
For language modeling task, we compare \textsc{LfD} with the following baseline models: \textbf{MLE}: uses the standard cross entropy in Eq.~\ref{eq::mle} for training. 
\textbf{Focal}~\cite{lin2017focal} downweights the loss of correctly-predicted tokens to deal with imbalance classification.
\textbf{UL}~\cite{welleck2019neural} penalizes the repetitive generation.

For dialogue generation task, in addition to \textbf{MLE} and \textbf{Focal}, following baselines are compared: \textbf{CP}~\cite{pereyra2017regularizing} regularizes the entropy of the model to alleviate over-confident predictions.
\textbf{FACE}~\cite{jiang2019improving} proposes to balance each token by considering their frequency in a training corpus. 
\textbf{Dialogue-UL}~\cite{li2019don} penalizes the overuse of frequently generated tokens using unlikelihood training. 
The implementation details of baseline models are described in Appendix~\ref{appendix::baselines}.

\noindent\paragraph{Evaluation Metrics}
For language modeling, we evaluate with the following metrics:
\textbf{Perplexity} to quantify the prediction difficulty of sequences by a model.
\textbf{Zipf Coefficient~(ZipC)}~\cite{holtzman2019curious} to measure the rank-frequency distribution of words in generated sequence.
\textbf{Repetition (Rep.)}~\cite{holtzman2019curious} to examine whether a sequence is stuck in repetitive loops.
\textbf{Unique~(Uniq.)}~\cite{welleck2019neural} to quantify the number of unique tokens in generated sequences.
\textbf{KL-Divergence~(KLD)}~\cite{csaky2019improving} to measure the divergence of unigram distributions between the generated texts and reference.

For dialogue generation, we use the following metrics:
\textbf{BLEU}~\cite{papineni-etal-2002-bleu} to measure the n-gram overlap between reference and generated sequences.
\textbf{Distinct}~\cite{li2016diversity} to calculate the ratio of unique N-grams among the generated sequences.
\textbf{self-BLEU}~\cite{selfbleu} to calculate the BLEU score of each sequence with other generated sequences.
Previously mentioned metrics~(\textbf{KLD}, \textbf{Context Overlap}, and \textbf{Source Entropy}) are also used. 
More details of each metric are available in Appendix~\ref{appendix::metrics_main}.

In abstractive summarization, we calculate the ratio of n-grams  in a summary that do not appear in a source article \textbf{(Novel-n)}~\cite{see2017get,narayan2018don}. 
We also measure the quality of generated summary with \textbf{Rouge}~\cite{banerjee-lavie-2005-meteor}.

\subsection{Main Results}
\noindent\paragraph{Language Modeling}
As shown in Table~\ref{app:lm_table}, our model shows similar token distribution with human-written texts in the corpus (KLD) and competitive performance with other models in PPL.
\textsc{LfD} significantly improves both Uniq. and ZipC, having minor gaps with the human texts. 
Surprisingly, \textsc{LfD} has a competitive result on Rep. with \textbf{UL} even in the lack of a penalty on repetition.

\noindent\paragraph{Dialogue Generation}
Results in Table~\ref{app:dialogue_table} show that \textsc{LfD} achieves the best scores in all metrics except for self-BLEU.
Interestingly, \textsc{LfD} shows the best BLEU scores, indicating that our approach can also contribute to increasing the similarity of generated responses with answer responses.
Although \textbf{FACE} shows better self-BLEU scores than \textsc{LfD}, its lower BLEU score may indicate that it fails to generate accurate response.

\begin{table*}[t!]
\centering
\begin{tabular}{ccccccc}
\toprule[0.3ex]

\textbf{Model}
 &\textbf{Rouge-1} &\textbf{Rouge-2} &  \textbf{Rouge-L}   & \textbf{Novel-1}(\%) & \textbf{Novel-2}(\%) & \textbf{Novel-3}(\%)  \\
\toprule
\textbf{MLE} & 40.64&	17.83&	37.67&		7.67&	20.00&	28.58 \\
\textbf{\textsc{LfD}} & 38.54&	15.48&		26.56&	8.59&	32.81	&48.16 \\
\textbf{Human}   & - & - & -  & 20.67 &56.46 &72.03 \\
\toprule
\end{tabular}

\caption{Evaluation results of abstractive summarzation task on CNN/DailyMail dataset. \textbf{Novel-N} indicates the ratio of novel N-gram in generated summaries.
}
\label{app:cnn_table}
\end{table*}

\noindent\paragraph{Abstractive Summarization}
\label{appendix::cnn}
We assume that the diversity of a summary is proportional to its \textit{abstractiveness}.
To measure the abstractiveness of summaries, we calculate the ratio of n-grams in a summary that do not appear in a source article~\cite{see2017get,narayan2018don}.
We also measure the quality of generated summary with ROUGE~\cite{banerjee-lavie-2005-meteor}.

As shown in Table~\ref{app:cnn_table}, the summaries generated by \textbf{MLE} contain fewer novel n-grams (i.e. low abstractivess) than human summaries.
\textsc{LfD} enhance the abstractiveness of generated summaries (+10.7\%, +64.1\%, and +68.5\% in Novel-1, Novel-2, and Novel-3 metrics, respectively), although the scores in ROUGE are slightly decreased~(e.g., -2.1 points in Rouge-1).
Based on these results, we confirm that the contribution of \textsc{LfD} is still valid in abstractive summarization which is aligned with empirical findings from \citet{goyal2022training}


\subsection{Amplifying \textit{De}generative Attributes}
We also evaluate the following models to confirm the validity of our framework:
\textbf{1) $f_{\theta_{D}}$}: We evaluate our \textit{De}generative model in Section~\ref{sec::degenerative} to check whether it actually captures negative behaviors and degenerate.
\textbf{2) LfD}$_{\text{MLE}}$: 
Instead of $f_{\theta_{D}}$ in Section \ref{sec::degenerative}, we use \textbf{MLE} as a \textit{De}generative model for PoE training. Results are shown in the bottom of Table~\ref{app:lm_table} and Table~\ref{app:dialogue_table}.
We first observe that $f_{\theta_{D}}$ performs significantly worse than other models, especially in diversity metrics, which implies that \textit{De}generative model successfully captures degenerative attributes in the dataset.
While \textbf{LfD}$_{\text{MLE}}$ also obtains performance gains compared with MLE in the diversity metrics, amplifying degenerative patterns mitigates degeneration and improves diversity more.

\section{Case Study}
\label{appendix::case_study}
We present generation examples for dialogue generation task in Table~\ref{table:response_examples}. As we can observe, our method usually increases the diversity of generated responses. For instance, given a dialogue context in the first example~(\textit{"Peter, how often do you exercise?"}), both MLE and Dialogue-UL models generate the responses with a generic phrase (\textit{“I’m not sure.”}). The FACE generation looks diverse at first glance, but it creates repeated n-gram (e.g., \textit{“after school every day"}, \textit{"day after school class"}). Finally, \textsc{LfD} creates a diverse and natural response by asking a question to the partner~(\textit{"How about you?"}).

\section{Conclusion}
In this paper, we discover that examples with low diversity are easy to be learned, leading models to degenerate in open-ended text generation tasks.
We propose \textsc{LfD}, a simple training approach that can be widely used to promote diversity without requiring specified negative behavior.
Experimental results on two representative tasks for open-ended generation confirm the validity and effectiveness of our approach.
\begin{table}[t!]
\small
\centering
\begin{tabular}{c|l}
\hline
\textbf{Context}    & \textbf{A}: Peter, how often do you exercise? \\ \hline
\textbf{MLE} &  I’m not sure. but I often go to the TV. \\ \hline
\textbf{FACE}        &  \begin{tabular}[c]{@{}l@{}} Not bad yet. How about taking photos out\\
every day after work at school every day\\
after school? They are pretty busy at school\\
every day after school every day after school\\
every day after school tomorrow evening\\
every day after school class day after school\end{tabular}  \\ \hline
\textbf{UL}        &  I’m not sure. I often go to a concert. \\ \hline
\textbf{Ours}        &  I’m pretty busy. How about you? \\ \hline
\textbf{Human} & Well, I swim and run every day. \\ \hline \hline

\textbf{Context}    & \begin{tabular}[c]{@{}l@{}} \textbf{A}: Room service.  \textbf{B}: Good morning . This \\is room 113 . I'd  like some breakfast, please.\end{tabular} \\
\hline
\textbf{MLE} &  Ok. How would you like to have your room? \\ \hline
\textbf{FACE}        &  \begin{tabular}[c]{@{}l@{}} Ok sir, here is some money left now and\\
would like some drinks or coffee beans.\\ Would you please fill out this form with us?
\end{tabular}  \\ \hline
\textbf{UL}        &  Ok, sir. How would you like to pay? \\ \hline
\textbf{Ours}        &  All right, sir. What would you like to order? \\ \hline
\textbf{Human} & Right. Excuse me. Mrs. Jones? \\ \hline

\end{tabular}
\caption{\label{table:response_examples} A generation example on DailyDialog dataset. UL denotes Dialogue-UL.}
\vspace{-0.1in}
\end{table}


\section*{Limitations}
In this work, we mainly investigate the relationship between the training of the generative model and the \textit{easiness} of undesired behavior that leads to degeneration.
For future work, we will extend our analysis of training dynamics into other degeneration problems such as hallucination or inconsistency, which are likely to be undesirable behaviors in other tasks.
Another limitation of LfD is that we focus on analyzing the learning dynamics of training examples in terms of the diversity.
Since the easily trained examples may consist of complex attributes more than low diversity, diminishing their impact on models may lead to an unintended generation.
In future work, we plan to conduct an in-depth analysis for easily trained examples to understand their characteristics.

\section*{Acknowledgements}
We thank Radhika Dua for the discussion and feedback on the paper.
This work was supported by the Institute of Information \& communications Technology Planning \& Evaluation (IITP) grant funded by the Korea government (MSIT) (No.2019-0-00075, Artificial Intelligence Graduate School Program (KAIST)), and the Institute of Information \& communications Technology Planning \& Evaluation (IITP) grant funded by the Korea government(MSIT) (No.2021-0-02068, Artificial Intelligence Innovation Hub).

\newpage
\bibliography{anthology,custom}
\bibliographystyle{acl_natbib}

\clearpage
\newpage

\appendix

\section*{Appendix}
\label{sec:appendix}

\section{Evaluation Metrics}
\label{appendix::metrics}
\subsection{Metrics in Preliminary Study}
\label{appendix::metrics_preliminary}
We describe further details in our evaluation metrics~\ref{sec::prelimiary_setup}.

\textbf{Context Overlap}~\cite{welleck2019neural}: We measure the ratio of shared bi-gram between contexts and responses as follows:
\begin{equation}
\begin{aligned}
& \mbox{Context Overlap}(x, y) =  \frac{ |N(x) \cap  N(y)|}{|N(y)|}
\end{aligned}    
\end{equation}
where $N(u)$ denotes the number of n-grams in utterance(s) $u$, while $x$ and $y$ indicate a dialog context and its response, respectively.

\textbf{Source Entropy}: We follow clustering-based method with MeanShift~\cite{comaniciu2002mean} algorithm to obtain entropy value of each response. We employ SimCSE-base~\cite{gao2021simcse} model finetuned on STS benchmark~\cite{cer2017semeval} to encode each text to a vector. The source entropy of a response is calculated as 

\begin{equation}
\begin{aligned}
& \mbox{H}_{\mbox{src.}}(c_y, C) = -\sum_{c_i \in C} p(c_i|c_y)\mbox{log}_{2} p(c_i|c_y)
\end{aligned}    
\end{equation}

where $C$ denotes the set of all clusters and $p(c_i|c_y)$ is the conditional probability of observing a dialog history from cluster $c_i$ given a response $y$ from cluster $c_y$.

\textbf{Repetition(Rep.)}:
We reinvent \textbf{Rep.} metric to compute repetitive patterns in ground-truth tokens inspired by the works~\cite{welleck2019neural,fu2021theoretical}.
The equation is as follow:
\begin{equation}
\mbox{Rep}(\textbf{x}) =  \frac{1}{\textbf{|x|}}\sum_{t=1}^{|\textbf{x}|} \textbf{I}[x_t \in x_{0:t-1}]
\label{eq::repetition}
\end{equation}

\subsection{Evaluation Metrics in Main Results}
\label{appendix::metrics_main}
\textbf{Perplexity}: To measure test perplexity in language modeling using decoder-only model, we regard condition $\text{x}$ as 50 prefixes and target $\text{y}$ as 100 of ground-truth next tokens.

\begin{equation}
\mbox{PPL}(\theta,\textbf{x},\textbf{y}) =  \frac{1}{\text{|y|}} \sum_{t=1}^{|\textbf{y}|} e^{-\log p(y_{t}|y_{<t},\textbf{x},\theta)}
\label{eq::ppl}
\end{equation}

\section{Implementation Details}
\label{appendix::details}
\subsection{Training Details}
In all experiments, we train language model on a single 3090 RTX GPU with 24GB of memory. We implemented all models with PyTorch using sentence-transformers library from UKPLab\footnote{https://github.com/UKPLab/sentence-transformers}.
In our experiments, we use 6-layer transformer decoder with GPT2~\cite{gpt2} tokenizer and 6-layer transformer encoder-decoder architectures with BERT-base-uncased~\cite{kenton2019bert} tokenizer for language modeling and dialogue generation tasks, respectively.
We choose the best checkpoints of models by using their validation loss.
We use Adam optimizer~\cite{kingma2015adam} with linear learning rate scheduler.
Learning rate is set to 1e-5 for language modeling task and 1e-4 for dialogue generation task.
The value of $\lambda$ that balances $\mathcal{L}_{\text{MLE}}$ and $\mathcal{L}_{\text{PoE}}$ is set to 0.25 and 0.5 for dialogue generation and language modeling task, respectively. 
We set the $R$ as 0.7 for both tasks, and set $K$ in Section~\ref{sec::degenerative} as the number of optimization steps for $f_{\theta}$ during 1 and 3 epochs on WikiText-103 and DailyDialog, respectively.
We set the $H$ in Section~\ref{sec::degenerative} as the number of optimization steps during an epoch on both tasks.

\subsection{Baseline Details}
\label{appendix::baselines}
We present more details of our baseline models.
We set the weight of repetition penalty in \textbf{UL} as 1.0.
The penalty weight of \textbf{Dialogue-UL} is set to 1000.
We set the $\gamma$ in \textbf{Focal} as 2.0. 
\textbf{Focal} aim to alleviate the negative effects of degenerative attributes by penalizing over-confident predictions of a model during training.
For \textbf{CP}, we set the weight of regularization term as 2.5 following the original paper.
For \textbf{FACE}, we use \textit{Output frequency} with \textit{Pre-weight} configurations for training.
The best checkpoint of \textbf{FACE} is chosen by using Distinct-1 metrics as suggested by the original paper.
In dialogue generation task, we finetune \textbf{CP}, \textbf{FACE}, \textbf{Dialogue-UL}, and \textsc{LfD} starting with \textbf{MLE}, and evaluate their performance in every 500 steps to find the best checkpoint.

\begin{table*}
\centering
\begin{subtable}{\textwidth}
\centering
\begin{tabular}{ccccccccc}
\toprule[0.3ex]
  \multicolumn{1}{c}{}& \multicolumn{2}{c}{\textbf{BLEU}} &    \multicolumn{2}{c}{\textbf{Distinct}} &\multicolumn{2}{c}{\textbf{self-BLEU}}  & \multicolumn{1}{c}{\textbf{KLD}} & \multicolumn{1}{c}{\textbf{Context}} \\ 
\multicolumn{1}{c}{\textbf{Model}} &  $n$=2 & $n$=3  &   $n$=2 & $n$=3  & $n$=3 & $n$=4 & &\textbf{Overlap}\\ \toprule \toprule

\textbf{MLE}     & 	18.70&	11.27&	   18.17&	36.26&		86.13&	77.32&	   0.98 & 13.16\\
\textbf{$\text{Dialogue-UL}^{\dagger}$}& 35.18&29.54 &  29.24&57.55 &  75.37&57.94 &  0.41 &  \textbf{9.45}	\\
\textbf{LfD}    & 	\textbf{35.24}&\textbf{29.72} &  \textbf{33.84}&\textbf{63.15}&  \textbf{71.31}&\textbf{51.68}&  \textbf{0.38}&  9.33\\
\textbf{Human}   & 	-    &	-    &		35.97&	67.00& 68.72&	47.34&-&     9.83 \\
\bottomrule
\end{tabular}
\end{subtable}

\caption{Evaluation results on DailyDialog with a pre-trained language model~(BERT2BERT). Greedy decoding algorithm is used for all models, following \citet{jiang2019improving}.
The indicators are the same as Table~\ref{app:dialogue_table}.
}
\label{app:pt_dialogue_table}
\vspace{-0.2cm}
\end{table*}

\begin{table}[t]

\begin{tabular}{c c l l c l}
\toprule[0.3ex]

\textbf{Model}
 &\textbf{PPL}  &  \textbf{ZipC}  & \textbf{Rep.} & \textbf{Uniq.}  \\
\toprule
\textbf{MLE}    
&\textbf{18.7}   & 1.16     & 1.3 & 8.72k  \\
\textbf{$\text{UL}^{\dagger}$}
&19.1  & \underline{0.95}      & \textbf{0.7} & \underline{9.22k}   \\

\textbf{Focal} 
&21.0  & 1.04     & 1.4 & 8.00k      \\
\textbf{LfD} 
&\underline{19.0} &   \bf{0.94}    & \underline{0.8} & \bf{9.50k}    \\
\textbf{Human}  
&-       & 0.93  & 0.2 & 10.9k \\ 



\toprule
\end{tabular}
\caption{Evaluation results on WikiText-103 with a pre-trained architecture. Top-$k$ sampling~\cite{fan2018hierarchical} is selected as decoding algorithm with $k$=20.
For \textbf{PPL}, the lowest score is the best performance.
}
\vspace{-0.3cm}
\label{app:pt_lm_table}
\end{table}

\subsection{Generation Details}
For open-ended text generation, we generate sequences for the evaluation by completing sequences from prefixes. Specifically, we preprocess test set of WikiText-103, select the first 50 tokens from each batch as prefixes, and lead models to generate a continuation of 100 tokens from the prefixes. 
We use top-\textit{k} sampling with $k$=20 as a decoding algorithm. 
For dialogue generation, we use a deterministic decoding algorithm (i.e. greedy decoding) following \citet{jiang2019improving}.

\subsection{Dataset Details}
\label{appendix::dataset}

\noindent\textbf{WikiText-103}
WikiText-103 contains 28.4k, 60, and 60 of articles on train, validation, and test split, respectively. We truncate sequence into 512 tokens in each example.

\noindent\textbf{DailyDialog}
DailyDialog dataset contains 13,118, 1000, and 1000 of multi-turn conversations on train, validation, and test split, respectively.
Following \citet{jiang2019improving}, we remove the dialogues with contexts or responses longer than 100 tokens to focus on short conversations.
This makes 55,404, 5130, and 4915 pair of dialog history and response in train, validation, and test split, respectively.

\section{Experiments with pre-trained language models}
\label{appendix::pretrain_result}
We also conduct experiments using pre-trained language models. For dialogue generation task, we use BERT2BERT architecture with BERT-base-uncased. For language modeling task, gpt2-small is used. Experimental results are shown in Table~\ref{app:pt_dialogue_table} and Table~\ref{app:pt_lm_table} for dialogue generation and language modeling tasks, respectively.

We first find that leveraging pre-trained models generally increase the overall performance of generation models. 
In dialogue generation task, \textsc{LfD} performs better than Dialogue-UL, a competitive baseline as shown in Table~\ref{app:dialogue_table}, except for Context Overlap scores. 
In language modeling task, our model usually performs better than other baselines. 
Based on these results, we confirm the validity of \textsc{LfD} even when they are applied with pre-trained language models.

\end{document}